\documentclass{article}

\usepackage[preprint]{corl_2024} % Use this for the initial submission.
\usepackage{amssymb}  
\usepackage{color}
\usepackage{xcolor}
\usepackage{amsmath}
\usepackage{listings}
\usepackage[position=bottom]{subfig}
\usepackage{tabularx,booktabs,caption,ragged2e}
\usepackage{graphics, wrapfig}
\usepackage[demo]{graphicx}
\newcolumntype{Y}{>{\RaggedRight\arraybackslash}X}

\lstdefinestyle{pythonstyle}{
    language=Python,
    basicstyle=\ttfamily\small,
    breaklines=false,
    commentstyle=\color{green!50!black},
    keywordstyle=\color{blue},
    stringstyle=\color{red},
    numbers=left,
    numberstyle=\tiny\color{gray},
    numbersep=5pt,
    frame=single,
    framesep=15pt,
    rulecolor=\color{black!30},
    showstringspaces=false,
    tabsize=4
}

\title{Agent-Arena: A General Framework for Evaluating Control Algorithms}

\author{
  Halid Abdulrahim Kadi\\
  University of St Andrews\\
  \texttt{ah390@st-andrews.ac.uk} 
  \And
  Kasim Terzi{\'c} \\
  University of St Andrews\\
  \texttt{kt54@st-andrews.ac.uk} \\
}

\begin{document}

\maketitle
%===============================================================================

\begin{abstract}
Robotic research is inherently challenging, requiring expertise in diverse environments and control algorithms. Adapting algorithms to new environments often poses significant difficulties, compounded by the need for extensive hyper-parameter tuning in data-driven methods. To address these challenges, we present Agent-Arena, a Python framework designed to streamline the integration, replication, development, and testing of decision-making policies across a wide range of benchmark environments. Unlike existing frameworks, Agent-Arena is uniquely generalised to support all types of control algorithms and is adaptable to both simulation and real-robot scenarios. Please see our GitHub repository \href{https://github.com/halid1020/agent-arena-v0}{https://github.com/halid1020/agent-arena-v0}.
\end{abstract}


%===============================================================================

\section{Introduction}

\begin{figure*}[t!]
    \centering
    \includegraphics[width=0.9\textwidth]{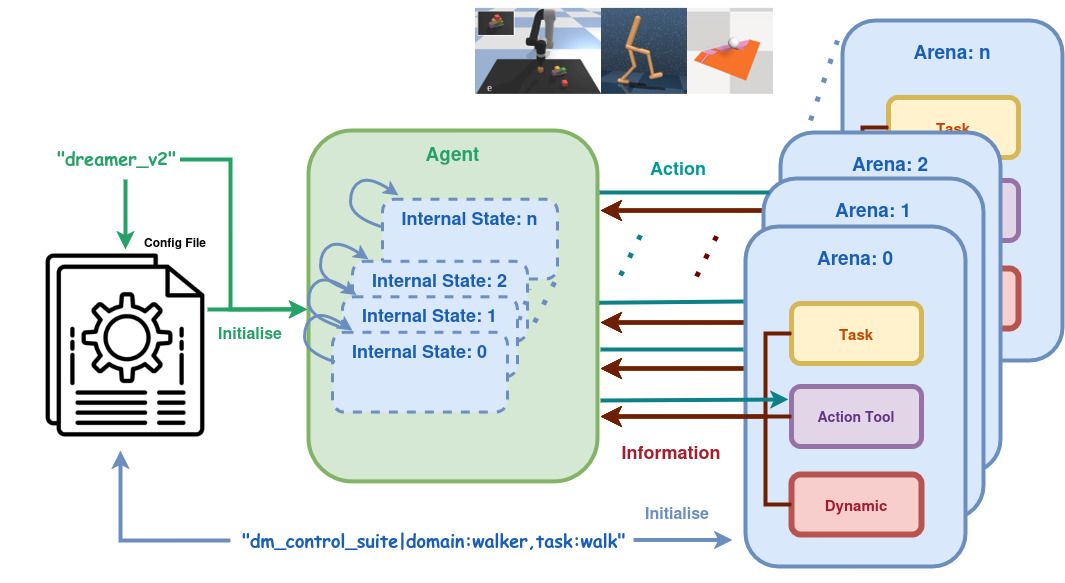} 
    \caption{\textit{Agent-Arena} Framework. Both the agent and a list of arenas can be initialised using agent and domain strings (represented by green and blue arrows), where the former is also constructed using its corresponding configuration files for data-driven controllers. The \texttt{Agent} (green block) maintains a set of internal states (blue block with a dashed boundary) for each of the created arenas (blue blocks with solid boundaries). It returns an appropriate set of actions (in cyan arrows) for the arenas after receiving information (in red arrows) from them. Upon construction, each arena has class variables for the intended task (yellow block), action tools (purple block), and underlying dynamics (red block). The information provided to the agents is aggregated from these three components. This framework currently supports the integration of deep-mind control suites \citep{tunyasuvunakool2020dmcontrol}, Raven \citep{zeng2021transporter}, and SoftGym \citep{lin2020softgym} benchmark environments. We use the former two for conducting sanity check on the baseline control algorithms used in this thesis, and the latter for the development of cloth-manipulation algorithms.} \label{fig:agent-arena}
\end{figure*}

Many robotic projects involve developing a control algorithm that activates hardware to achieve intended tasks in real physical environments after processing sensor inputs. In modern robotics, an increasing number of controllers are first developed and tested in simulation before being transferred and deployed to real-world scenarios. This process requires the same control algorithm to be adaptable across different domains.

With the advent of deep learning (DL), both the perception and control components of robotic manipulation systems (RMSs) are gradually being replaced by neural-network (\text{NN})-based methods, introducing new challenges to the field. These challenges include fine-tuning a large set of hyper-parameters while ensuring wider-context generalisability in practice. Over the past half-decade, an increasing number of general-purpose deep controllers \citep{hafner2019learning, zeng2021transporter, hafner2023mastering, chi2023diffusionpolicy, black2024pi0} have been proposed. Meanwhile, many of these methods are developed and tested on different types of tasks and benchmark domains. The community currently lacks a general framework that enables convenient comparison and evaluation of classical and data-driven, as well as general and domain-specific, controllers across various domains.

To benefit the robot-learning and wider robotics community, we propose \textit{Agent-Arena}, a general Python framework that provides an interface for controllers and environments to integrate, develop, examine, and compare various types of controllers. In this framework, an agent performs actions in an arena, which then returns state information to the agent in a closed-loop manner. Although we mainly use this framework for cloth-manipulation tasks in this thesis, it can be utilised to conduct research in wider range of robotic manipulation tasks.

Integrating \textit{SoftGym} \citep{lin2020softgym}, \textit{Raven} \citep{zeng2021transporter},  \textit{DeepMind Control Suite} \citep{tunyasuvunakool2020dmcontrol} simulation benchmark environments as well as \textit{ROS1 Noetic} and \textit{ROS2 Humble} for real-world configurations, \textit{Agent-Arena} is the basis for many published projects \citep{kadi2024planet, kadi2024mjtn, kadi2025draper}. It adopts \textit{OpenAI Gym} \citep{openai2016gym} for action configuration, \textit{Python Ray} \citep{moritz2018ray} for multi-processing, and \textit{Python Zarr} \citep{zarr} for implementing general trajectory dataset. This chapter introduces the basic structure of this framework, along with simple example scripts for users.

An agent can be initialised using a configuration file and its name. It can represent either a domain-specific or a general method, and it may be an analytical or data-driven approach. An arena, on the other hand, can be initialised using a Markov Decision Process (\text{MDP}) string and supports a wide range of simulated task environments, including those provided by \textit{MuJoCo} \citep{todorov2021mujoco}, \textit{PyBullet} \citep{coumans2021pybullet}, \textit{Atari} \citep{bellemare2013arcade}, and \textit{NVIDIA Flex} \citep{lin2020softgym, macklin2014unified}. Additionally, we provide training and evaluation tools along with APIs that incorporate adapter and builder patterns to simplify and accelerate the development of robotic and decision-making policies.

We adopt the term \textit{Agent-Arena} from renowned cognitive scientist Prof John Vervaeke \citep{vervaeke2019awakening}, who addresses the "meaning crisis" of the modern world by re-engineering our understanding of cognition, wisdom, and self-transcendence within a scientific framework. We build this framework for the general control domain from a first-principles perspective, enabling the integration of many different controllers and benchmark environments. In summary, we make the following  contributions:

1. We develop a multi-agent, multi-arena interaction framework, where single or multiple agents interact with multiple task-augmented environments, i.e., arenas, in parallel.

2. We provide a uniform dataset layout for control tasks, which supports both trajectory and cross-trajectory sampling with adjustable sequence lengths, while offering configurable observation and action outputs.

3. We implement a set of visualisation and data logging utilities to visualise and store various types of robotic data.

4. We integrate with \textit{ROS1 Noetic} and \textit{ROS2 Humble}, facilitating the seamless transfer of simulation-trained policies to real-world robotic systems.

\section{Background} \label{sec:agent-arena-related-work}
Many frameworks proposed by the robotic and reinforcement learning (\text{RL}) community try to provide a general paradigm for facilitating Embodied AI research, such as \textit{Embodied} \citep{hafner-embodied}, \textit{Dopamine} \citep{castro2018dopamine},\textit{OpenAI Gym} \citep{openai2016gym}, \textit{AllenAct} \citep{AllenAct}, and \textit{PyRoboLearn} \citep{delhaisse2020pyrobolearn}.

\textit{Embodied} \citep{hafner-embodied} has been used to develop general deep reinforcement learning (\text{DRL}) algorithms, such as \textit{DayDreamer} \citep{wu2022daydreamer} and \textit{Dreamer V3} \citep{hafner2023mastering}, for doing research in both simulation and real-world; it integrates the visual control domains such as \textit{DeepMind Control Suite} \citep{tunyasuvunakool2020dmcontrol}, \textit{DeepMind Lab} \citep{beattie2016deepmind}, \textit{MineRL} \citep{milani2020minerl} and \textit{Atari 2600, ALE} \citep{bellemare2013arcade}. However, this framework does not support the integration of imitation learning (\text{IL}) and classical control methods. \textit{Dopamine} \citep{castro2018dopamine} is another framework for developing \text{DRL} but restricted on nueral network (\text{NN}) package \textit{Tensorflow} and \textit{Atari 2600, ALE} \citep{bellemare2013arcade} environments.

\textit{AllenAct} \citep{AllenAct}, on the other hand, supports the development of both \text{DRL} and \text{DIL} methods on more complex and real-life environments, such as \textit{iTHOR} \citep{kolve2017ai2}, \textit{RoboTHOR} \citep{deitke2020robothor}, and \textit{Habitat} \citep{savva2019habitat}, along with the environments in \textit{OpenAI Gym} \citep{openai2016gym} and \textit{MiniGrid} \citep{MinigridMiniworld23}. It provides a wholesome tutorial and many integrated methods for training and evaluations, but their framework only provides interface for learning-based methods without considering classical control methods. \textit{AllenAct} disentangles the task from the dynamics and lets tasks provide the action to the agents. In contrast, in our framework \textit{Agent-Arena}, we further decouple tasks and actions so that various different actions can apply to the same task and vice versa; as a result, an arena will be a encapsulation of a dynamic, action tool and a task in our framework.

\textit{OpenAI Gym} \citep{openai2016gym} adopts the structure of Partially Observable Markov Decision Process \text{(POMDP)} to provide a standardised interface for integrating or creating new environments with the main intention to develop \text{RL} controllers, and the action class is expected to wrap around the underlying environment. It provides the integration for \textit{Atari} \citep{bellemare2013arcade}, \textit{Mujoco} \citep{todorov2021mujoco}, and some other simple game environments. Moreover, it provides a nice interface for observation and action space, and \textit{AllenAct} also adopts it for their own framework. We also use their action space in our framework, but the one major difference of our framework compared to \textit{OpenAI Gym} is that, the action tool is a class variable of an arena instead of a wrapper.

\textit{PyRoboLearn} \citep{delhaisse2020pyrobolearn} is another framework that is specifically designed for developing robot learning  methods with over 60 different robots in \textit{Pybullet} simulator \citep{coumans2021pybullet}. There are also many other benchmark environments that are designed for specific purpose. \textit{RLBench} \citep{james2020rlbench} provides a set of single-arm robotic manipulation task for benchmarking \text{RL} algorithms. \textit{CARLA} \citep{dosovitskiy2017carla} is an open-source simulator for urban autonomous driving, and it supports developing and evaluating driving models. Having a similar name as our framework, \textit{Arena} \citep{song2020arena} is a general evaluation framework that support multi-agent training environments. There are also many other benchmark environments for developing \text{DRL} algorithms, such as \textit{ViZDoom} \citep{kempka2016vizdoom}, and \textit{Sapien} \citep{xiang2020sapien}, and for facilitating robot learning, such as \textit{Gibson Env} \citep{xia2018gibson}.

In summary, \textit{Agent-Arena} is the only general framework supports the development and evaluation of all kinds of control algorithms in both simulation and real-world environments. The development of this framework is essential for facilitating and accelerating the accomplishment of this thesis.

\section{Framework}

\textit{Agent-Arena} is a general framework for running controllers in different environments in parallel in a closed-loop manner. The \texttt{Agent} supports many general-purpose reinforcement learning (\text{RL}) and imitation learning (\text{IL}) algorithms, as well as domain-specific control approaches; it also supports both analytical and data-driven methods. \texttt{Arena} adopts adapter and builder patterns to help users to initialise various types of environments easily through a domain string. Not only \textit{Agent-Arena} supports \text{POMDP} and \text{MDP} control frameworks, it underpins all possible modelling of control problems with a closed information-action loop. The hyper-parameters of the integrated controllers are stored in a \texttt{yaml} configuration files for replication of the tested results in this thesis. Please see the architecture of this framework in Figure \ref{fig:agent-arena}, 

Program \ref{lst:example-code} is an example client program to run an training experiment in the framework using an already integrated algorithm \texttt{dreamer\_v2} and \texttt{dm\_control\_suit} benchmark environments. In this example, the script loads the provided configuration file \texttt{default} for the agent-arena string pair; however, users are also allowed to create their own customised configuration in the type of \texttt{DotMap} and provides all the required parameters.

\begin{lstlisting}[style=pythonstyle,basicstyle=\tiny\ttfamily,caption={Example Python Code for Training and Evaluating an Agent},label={lst:example-code}]
import agent_arena as agar

arena_name = "dm_control_suite|domain:walker,task:walk"
agent_name = "dreamer_v2"
save_dir = "./results/dreamer_on_walker-walk"

# Load the configuration file as type of DotMap stored 
# as "default.yaml" for the (agent_name, arena_name) pair.
config = agar.retrieve_config(
    agent_name, arena_name, config_name="default")
    
arena = agar.build_arena(arena_name)
agent = agar.build_agent(agent_name, config=config)

arena.set_log_dir(save_dir)
agent.set_log_dir(save_dir)

agar.train_and_evaluate(agent, arena)
\end{lstlisting}

The application programming interface (API) of this framework includes the \texttt{Agent}, \texttt{TrainableAgent}, \texttt{Arena}, \texttt{Task}, \texttt{ActionTool}, \texttt{TrajectoryDataset}, and \texttt{Logger} classes. These classes provide an interface and a template for developing customised instances for user-specific usage. The framework also provides \texttt{build\_agent}, \texttt{build\_arena}, \texttt{retrieve\_config}, and \texttt{build\_transform} functions for initialising built-in agents and arenas to replicate experiments. In addition, it offers the \texttt{run}, \texttt{evaluate}, \texttt{validate}, and \texttt{train\_and\_evaluate} functions for running the agent in a single trial of the arena, evaluating it on the arena's test trials, validating it on the arena's validation trials, and training the agent before evaluating it at the end while performing intermediate validations. Furthermore, it provides the \texttt{perform\_single} and \texttt{perform\_parallel} functions to run the agent on a single arena or across multiple arenas in parallel. We define the types of major communication variables as inf Program \ref{lst:types} under \texttt{utilities/types.py}.

All transitional information is returned in the form of map/dictionary to the agents. Agent and Arena are completely decoupled, except for Oracle agents that can get access to the arena field of the \texttt{Information} to get access to the true dynamical state of the environment. Both agent and arena have their own individual loggers that are independent from each other. The arena's logger stores how action interacts with the environment, the agent's logger records the agent's step-wise internal states. This requires the agent to be aware of which arena it is interacting with; hence when feeding a list of information to the \texttt{act} function, the information should include arena id that is allocated for parallelisation purpose; so that the agent would be able update the internal state for the corresponding episode.

We also provide a set of visualisation and storing functions for depth, rgb, and mask images as well point-clouds and videos in \texttt{utilities/visual\_utils.py}. We provide many testing scripts to ensure the sanity of the provided functionalities. Users can also run the customised examples in \texttt{example} folder to learn and run the framework. Now, we are going to dive deep into the provided classes in details.

\newpage
\begin{lstlisting}[style=pythonstyle,basicstyle=\tiny\ttfamily,caption={Typings in Agent-Arena},label={lst:types}]
from gym.spaces import Space
from typing import Dict, Any

ActionSpaceType = Space
ActionType = Any
InformationType = Dict[str, Any]
ArenaIdType = int
ActionPhaseType = str
\end{lstlisting}

\subsection{Agent}

The \texttt{Agent} class serves as a base class for all agents,
 encompassing all types of control methods,  It includes the \texttt{reset}, \texttt{act}, \texttt{init}, and \texttt{update} methods to interact with an arena in a closed-loop manner, where the latter two methods are reserved for agents to update their internal states. These methods are arena-sensitive, meaning they recognise the source arena of the received information and execute internal operations accordingly. The \texttt{Agent} class also provides the \texttt{success}, \texttt{terminate}, and \texttt{get\_phase} methods to check whether the agent has successfully completed the task and to retrieve the current phase of the task from the agent's perspective. These methods return results for all operating arenas. Note that an open-loop controller can maintain an internal step counter to enable long-horizon action outputs. Please refer to Appendix \ref{sec:agent-arena-interface-agent} for further details on its interface.

\paragraph{TrainableAgent}

Extending from \texttt{Agent}, \texttt{TrainableAgent} is designed for learning-based methods, such as \text{RL} and \text{IL} algorithms. It provides functionalities to save and load training checkpoints from either a default or a specified directory. Additionally, it offers a \texttt{train} function that can take a list of arenas as an input argument in cases where it needs to interact with the arenas during the training process. Please refer to Appendix \ref{sec:agent-arena-interface-trainableagent} for further details on its interface. Note that the way the actions are returned by an agent should be defined in the \texttt{yaml} configuration files for the specific arenas they interact with, unless the agent is developed specifically for a certain arena. Furthermore, we use the \texttt{update\_steps} variable for saving checkpoints instead of \texttt{epochs}, as the former is more general, whereas the latter refers to updating the model by traversing every data point in a fixed dataset once.

\subsection{Arena}
The interface of the \texttt{Arena} is quite challenging to design, and it has undergone many structural improvements throughout the study. As humans, acting as agents, we only interact with the dynamics of our own universe; there are no objectively specified task goals (other than survival) and no predefined action primitives. In robotics, action primitives are often hierarchical, meaning that an arena does not need to adhere strictly to the lowest fundamental levels of control. Therefore, our framework should be capable of adapting to higher-level action abstractions, such as moving towards specified waypoints and \text{PnP}. One of the key design questions is how we should integrate goals and tasks with an environment's dynamics. We decided to make these elements variables of an \textit{Arena}, and we designed an arena to be generally initialised by specified domains, tasks, and action primitives, each corresponding to a specific class.

\texttt{Task} and \texttt{ActionTool} are independent of the \texttt{Arena}, but the \texttt{Arena} needs to have them in order to transition and evaluate. The \texttt{Task} and \texttt{ActionTool} classes provide associated services by passing the arena as a method argument. Be aware that many developed benchmark environments come with their own predefined action primitives and tasks; in these cases, re-engineering these benchmarks to fit exactly into our framework would be arduous. Thus, we also provide flexibility to integrate these benchmark environments easily. If the action space is not specified in the initialisation string, the arena inherits the action space from the wrapped domain provided by the underlying benchmark environment; otherwise, it follows the lowest level of action abstraction. Similarly, if the task is not specified, an arena inherits the task from the wrapped domain provided by the underlying benchmark environment; otherwise, it operates without goals and/or rewards. Please refer to Appendix \ref{sec:agent-arena-interface-arena} for further understanding of the \texttt{Arena} interface.

\paragraph{Task}
As discussed above, the \texttt{task} of an arena is decoupled from it so that the \texttt{task} can be applied to multiple arenas with different physical engines. Also, the same arena can be used to perform different types of tasks. In addition, one can specify goals with different data types for achieving a task; one can provide an image, a configuration for the whole or partial state of the environment, a piece of text, a demonstrated trajectory, as well as a mixture of all these. Hence, we decided to let the goal be of type dictionary. The agent then decides which types it will use to finish the task. Please refer to Appendix \ref{sec:agent-arena-interface-task} for further understanding of its interface.

\paragraph{Action Tool}
Similarly, the \texttt{ActionTool} class is also decoupled from \texttt{Arena} for a similar purpose as \texttt{Task}. Hence, an action tool can be applied to different simulation-based arenas, and each arena can take different types of action primitives for achieving the same task under the same physics.Please refer to Appendix \ref{sec:agent-arena-interface-actiontool} for further understanding of its interface.

\paragraph{Multiprocessing}
Running an agent sequentially on a single arena can be inefficient for training and evaluating learning-based controllers. Although we can run multiple instances of the same programme to achieve multi-processing, one has to do so manually. Multi-processing is especially important for collecting data and evaluating agents in parallel on their test trials to save time. Inspired by \citep{canberk2023cloth}, we utilise the Python \textit{Ray} \citep{moritz2018ray} package to achieve this purpose. We assign each instance of an arena a unique ID and treat each arena as a \texttt{Ray Actor}. We also provide \texttt{utilities/perform\_parallel.py}, which can run single or multiple agents on a list of arenas with the target episode configurations from start to finish whilst collecting step-wise information of the arenas and internal states of the agents. Usually, we want to avoid having multiple agents for multiple arenas, because the neural networks of a trainable agent can be large, and we should be able to take advantage of parallelisation of the networks for interacting with multiple arenas simultaneously.

\subsection{Trajectory Dataset}

We create a general trajectory dataset for control problems that allows a client to add and access transitional data quickly and easily. We define it as a PyTorch \texttt{Dataset} under \texttt{utilities/trajectory\_dataset.py}, and wrap it around the Python \texttt{zarr} package \citep{zarr}. \texttt{zarr} is a format for the storage of chunked, compressed, N-dimensional arrays; it organises arrays into hierarchies via groups and supports reading and writing an array concurrently from multiple threads or processes. These properties are ideal for saving hierarchical information and action data structures of our framework, and it supports a reasonable loading speed from storage for training learning-based agents.

We enable the dataset to return step-wise transitional data, trial-wise trajectory data, and cross-trial sequential data depending on the user's preference. The storage hierarchy of the dataset is defined by \texttt{obs\_config}, \texttt{act\_config} and \texttt{goal\_config}, which are expected to specify the saved names, storage shapes, and return names for these data types. We also support saving from the last interrupted checkpoint by appending to the output file. This dataset also supports splitting the stored data into training, validation, and evaluation partitions with specified ratios. Program \ref{lst:data-collect} shows an example for collecting data using our general-purpose dataset in \textit{Agent-Arena}.

\newpage

\begin{lstlisting}[style=pythonstyle,basicstyle=\tiny\ttfamily,caption={An Example for Collecting Data using Agent-Arena},label={lst:data-collect}]
import numpy as np
import cv2
import agent_arena as ag_ar
from agent_arena.utilities.trajectory_dataset import TrajectoryDataset
from agent_arena.utilities.perform_single import perform_single

obs_config = {'rgb': {'shape': (128, 128, 3),  'output_key': 'rgb'}}
act_config = {'norm-pixel-pick-and-place': {'shape': (2, 2), 'output_key': 'default'}}

total_trials = 5
data_path = 'exmaple_trajectory_data'
agent_name = "oracle-garment|mask-biased-pick-and-place"
arena_name = "softgym|domain:mono-square-fabric,initial:crumpled," + \
       "action:pixel-pick-and-place(1),task:flattening"

config = ag_ar.retrieve_config(agent_name, arena_name, "")
arena = ag_ar.build_arena(arena_name + ',disp:False')
agent = ag_ar.build_agent(agent_name, config)

dataset = TrajectoryDataset(
    data_path=data_path, io_mode='a', whole_trajectory=True,
    obs_config=obs_config, act_config=act_config)

while dataset.num_trajectories() < total_trials:
    res = perform_single(arena, agent, mode='train', max_steps=3)
    observations, actions = {}, {}
    for k, v in obs_config.items():
        obs_data = [info['observation'][k] for info in res['information']]
        observations[k] = [cv2.resize(obs.astype(np.float32), v['shape'][:2]) for obs in obs_data]
    for k, v in act_config.items():
        act_data = [a[k] for a in res['actions']]
        actions[k] = [np.stack([act['pick_0'], act['place_0']]).reshape(*v['shape']) \
            for act in act_data]

    dataset.add_trajectory(observations, actions)


\end{lstlisting}

\section{Conclusion}

\textit{Agent-Arena} is the Python framework developed to assist roboticist in conducting experiments quickly and easily. It is the basis of numerous robotics projects \citep{kadi2024mjtn, kadi2024planet, kadi2025draper}. This framework can be installed on the most modern Ubuntu machines using an \texttt{anaconda} environment. We also managed to achieve integration with \textit{ROS1 Noetic} and \textit{ROS2 Humble} for conducting real-world experiments on \textit{Panda} and \textit{UR3e} robot arms. Nevertheless, there are still some important features that we would like to develop in the future.

\subsection*{Limitations and Future Work}
\textit{Agent-Arena} is a powerful tool that sufficiently facilitates several robotics projects \citep{kadi2024mjtn, kadi2024planet, kadi2025draper}; however, we can extend its functionalities and resolve following limitation in the future.

\begin{enumerate}
    \item The current Agent-Arena framework is constrained to single-machine operations for training and evaluation. We aim to expand its capabilities by developing functionalities that enable distributed computing, allowing agents to run and train in parallel across multiple machines in a cluster environment.
    \item We recognise the need to decouple environmental sensing from dynamics, enabling diverse observation setups within the same dynamic-task configuration. To achieve this, we plan to separate sensing tools from the arena, mirroring our approach with action tools and tasks.
    \item The \texttt{TrajectoryDataset} currently loads data from storage without considering the size of stored data. We intend to implement an adaptive loading mechanism that efficiently utilises available memory, transferring all data into RAM when sufficient resources are detected.
    \item We will enhance the framework's versatility by integrating a broader range of general-purpose algorithms and benchmark environments.
    \item We are committed to developing advanced functionalities for transfer learning and multi-task learning, specifically tailored for robotic applications, thereby expanding the framework's applicability in complex scenarios.
\end{enumerate}

\newpage
\bibliography{example}

\newpage
\label{apx:agent-arena}

\section{Agent} \label{sec:agent-arena-interface-agent}

\begin{lstlisting}[style=pythonstyle,basicstyle=\tiny\ttfamily,caption={Agent Interface in Agent-Arena},label={lst:agent}]
from abc import ABC, abstractmethod
from typing import Dict, List, Any
from dotmap import DotMap

from ..utilities.logger.dummy_logger import DummyLogger
from ..utilities.types import ActionType, InformationType, \
    ArenaIdType, ActionPhaseType

class Agent(ABC):
    def __init__(self, config: DotMap):
        self.config: DotMap = config
        self.name: str = "agent"
        self.internal_states: Dict[ArenaIdType, InformationType] = {}
        self.logger = DummyLogger()

    def get_name(self) -> str:
        """Return the name of the agent. 
        This will be used but not limited for logging."""
        return self.name

    def set_log_dir(self, logger: Any) -> None:
        """Set the log directory for the agent."""
        self.logger.set_log_dir(logger)

    def reset(self, arena_ids: List[ArenaIdType]) -> List[bool]:
        """
        Reset the agent before a new trial for the given arena_ids.
        
        Args:
            arena_ids: List of arena identifiers.
        
        Returns:
            A list of booleans indicating if 
            the resets were successful for each arena.
        """
        for arena_id in arena_ids:
            self.internal_states[arena_id] = {}
        return [True for _ in arena_ids]

    def init(self, informations: List[Dict[str, Any]]) -> List[bool]:
        """
        Initialise the agent's internal state given the initial information.
        
        Args:
            informations: Initial information for the agent from the reset arenas.
        
        Returns:
            A list indicating if the initialization was successful.
        """
        return [True for _ in informations]

    def update(self, informations: List[InformationType], actions: List[ActionType]) -> List[bool]:
        """
        Update the agent's internal state given the current information and action.
        
        Args:
            informations: Current list of informations 
                          for the agent from the arenas.
            actions: Actions taken by the agent.
        
        Returns:
            A list indicating if the update was successful.
        """
        return [True for _ in informations]

    @abstractmethod
    def act(self, informations: List[InformationType], update: bool = False) -> List[ActionType]:
        """
        Produce actions given the current informations from the arena, 
        update the internal state if required.
        
        Args:
            informations: Current information for the agent.
            update: Whether to update the agent's internal state; 
                    do not update if called `update` 
                    and/or 'init' method before.
        
        Returns:
            A list containing the agent's action.
        """
        raise NotImplementedError

    def success(self) -> Dict[ArenaIdType, bool]:
        """Check if the agent thinks it succeeded in each arena."""
        return {arena_id: False for arena_id in self.internal_states.keys()}

    def terminate(self) -> Dict[ArenaIdType, bool]:
        """Check if the agent thinks it should terminate in each arena."""
        return {arena_id: False for arena_id in self.internal_states.keys()}

    def get_phase(self) -> Dict[ArenaIdType, ActionPhaseType]:
        """Get the current action phase for each arena."""
        return {arena_id: 'none' for arena_id in self.internal_states.keys()}

    def get_state(self) -> Dict[ArenaIdType, InformationType]:
        """
        Return intermediate state after applying act, init or update method.
        
        Returns:
            A dictionary containing the internal states for each arena.
        """
        return self.internal_states
\end{lstlisting}

\newpage
\section{TrainableAgent} \label{sec:agent-arena-interface-trainableagent}

\begin{lstlisting}[style=pythonstyle,basicstyle=\tiny\ttfamily,caption={TrainableAgent Interface in Agent-Arena},label={lst:trainable-agent}]
from abc import abstractmethod
from typing import Optional, List
from ..arena.arena import Arena
from ..utilities.utils import TrainWriter
from .agent import Agent

class TrainableAgent(Agent):
    def __init__(self, config):
        super().__init__(config)
        self.name: str = "trainable-agent"
        self.train_writer: TrainWriter = TrainWriter()

    def train(self, update_steps: int, arenas: Optional[List[Arena]] = None) -> bool:
        """
        Train the agent, optionally on the provided arenas.

        Args:
            update_steps: Number of update steps to perform.
            arenas: Optional list of arenas to train on.

        Returns:
            bool: True if the training is successful, False otherwise.
        """
        return False

    def load(self, path: Optional[str] = None) -> int:
        """
        Load the latest agent checkpoint from the specified path or the logger's log directory.

        Args:
            path: Optional path to load the checkpoint from.

        Returns:
            int: The checkpoint number that was loaded, or -1 if loading was unsuccessful.
        """
        return -1

    def load_checkpoint(self, checkpoint: int) -> bool:
        """
        Load the agent from a specific checkpoint in its logger's log directory.

        Args:
            checkpoint: The checkpoint number to load.

        Returns:
            bool: True if the loading is successful, False otherwise.
        """
        return False

    @abstractmethod
    def save(self, path: Optional[str] = None) -> bool:
        """
        Save the current agent checkpoint to the specified path.

        Args:
            path: Optional path to save the checkpoint to, 
                  if path is None, save to the logger's log directory.

        Returns:
            bool: True if the saving is successful, False otherwise.
        """
        raise NotImplementedError

    @abstractmethod
    def set_train(self) -> None:
        """Set the agent to training mode."""
        raise NotImplementedError

    @abstractmethod
    def set_eval(self) -> None:
        """Set the agent to evaluation mode."""
        raise NotImplementedError

    def get_train_writer(self) -> TrainWriter:
        """
        Get the writer for logging training data.

        Returns:
            TrainWriter: The train writer object.
        """
        return self.train_writer
\end{lstlisting}

\newpage
\section{Arena} \label{sec:agent-arena-interface-arena}

\begin{lstlisting}[style=pythonstyle,basicstyle=\tiny\ttfamily,caption={Arena Interface in Agent-Arena},label={lst:arena}]
from abc import ABC, abstractmethod
from typing import Dict, Any, List, Optional
import numpy as np

from ..utilities.logger.dummy_logger import DummyLogger
from ..utilities.types import ActionType, InformationType, ActionSpaceType
from .dummy_task import DummyTask
 
class Arena(ABC):
    """
        Abstract class for defining an arena in a control problem.
    """

    def __init__(self):
        self.name = "arena"
        self.mode = "train"
        self.setup_ray(id=0)
        self.disp = False
        self.random_reset = True
        self.logger = DummyLogger()
        self.task = DummyTask()

    def set_log_dir(self, path: str):
        """
        Set the log directory for the logger.

        Args:
            path: The path to the log directory.
        """
        self.logger.set_log_dir(path)

    ##### The following is used by api #####
    def get_name(self) -> str:
        """
        Get the name of the arena.

        Returns:
            str: The name of the arena.
        """

        return self.name
    
    def set_disp(self, flg: bool):
        """
        Set the display flag for GUI demonstration.

        Args:
            flg (bool): True to enable display, False to disable.
        """

        self.disp = flg

    
    @abstractmethod
    def get_eval_configs(self) -> List[Dict[str, Any]]:
        """
        Get configurations for evaluation episodes.

        Returns:
            List of configurations for evaluation episodes.
        """
        raise NotImplementedError
    
    @abstractmethod
    def get_val_configs(self) -> List[Dict[str, Any]]:
        """
        Get configurations for validation episodes.

        Returns:
            List of configurations for validation episodes.
        """
        raise NotImplementedError

   
    # Core arena methods
    @abstractmethod
    def reset(self, episode_config: Optional[Dict[str, Any]] = None) -> InformationType:
        """
        Reset the arena for a new trial.

        Args:
            episode_config (Optional[Dict[str, Any]]): Configuration for the episode.
            
            if episode_config is None, 
                it set episode_config to {'eid': <random>, 'save_video': False}
            
            if eid is not given in episode_config and `random_reset` is True, 
                then sample a random eid;

        Returns:
            InformationType: Information about the arena state after reset.
        """
        
        raise NotImplementedError
    
    @abstractmethod
    def step(self, action: ActionType) -> InformationType:
        """
            This method take an `action` to the environment, 
            and return `information` about the arena.
        """
        raise NotImplementedError

    @abstractmethod
    def get_frames(self) -> List[np.ndarray]:
        """
        Get the list of frames collected so far.

        Returns:
            List[np.ndarray]: List of frames.
        """
        raise NotImplementedError
    
    @abstractmethod
    def clear_frames(self):
        """
        Clear the list of collected frames.
        """
        raise NotImplementedError
    
    @abstractmethod
    def get_goal(self) -> InformationType:
        """
        Get the goal of the current episode.

        Returns:
            InformationType: Information about the current goal.
        """
        raise NotImplementedError
    
    @abstractmethod
    def get_action_space() -> ActionSpaceType:
        """
        Get the action space of the arena.

        Returns:
            ActionSpaceType: The action space defined using gym.spaces.
        """
        raise NotImplementedError
    
    @abstractmethod
    def sample_random_action():
        """
        Sample a random action from the action space.

        Returns:
            A uniformly sampled action from the action space.
        """
        raise NotImplementedError
    
    def set_train(self):
        """
        Set the arena to sample only training episodes.
        """
        self.mode = "train"

    def set_eval(self):
        """
        Set the arena to sample only evaluation episodes.
        """
        self.mode = "eval"

    def set_val(self):
        """
        Set the arena to sample only validation episodes.
        """
        self.mode = "val"
    
    @abstractmethod
    def get_no_op(self) -> ActionType:
        """
        Get the no-op action (action with no effect on the environment).

        Returns:
            ActionType: The no-op action.
        """
        raise NotImplementedError
    
    def evaluate(self) -> Dict[str, Any]:
        """
        Evaluate the arena and return metrics.

        Returns:
            Dict[str, Any]: A dictionary of evaluated metrics.
        """
        return self.task.evaluate(self, metrics={})
    
    @abstractmethod
    def get_action_horizon(self) -> int:
        """
        Get the action horizon (length of an episode) of the arena.

        Returns:
            int: The action horizon.
        """
        raise NotImplementedError
    
    def get_num_episodes(self) -> int:
        """
        Get the number of possible episodes in the arena under the current mode.

        Returns:
            int: The number of possible episodes, or -1 if undefined.
        """
        return -1

    def set_task(self, task):
        """
        Set the task for the arena.

        Args:
            task: The task to be set.
        """
        self.task = task
    
    def success(self):
        return self.task.success(self)
    
    def setup_ray(self, id):
        """
            This method sets up the ray handle for the arena for multi-processing.
        """
        self.id = id
        self.ray_handle = {"val": id}
\end{lstlisting}

\newpage
\section{Task} \label{sec:agent-arena-interface-task}
\begin{lstlisting}[style=pythonstyle,basicstyle=\tiny\ttfamily,caption={Task Interface in Agent-Arena},label={lst:task}]
from abc import ABC, abstractmethod
from typing import Any, Dict, List
from ..utilities.types import InformationType
from .arena import Arena

class Task(ABC):
    """
    Abstract base class defining the interface for tasks in an arena.
    """

    @staticmethod
    @abstractmethod
    def reset(arena: Arena) -> InformationType:
        """
        Reset the task for a new episode.

        Args:
            arena: The arena in which the task is being performed.
        """
        raise NotImplementedError

    @staticmethod
    @abstractmethod
    def success(arena: Arena) -> bool:
        """
        Check if the task has been successfully completed.

        Args:
            arena: The arena in which the task is being performed.

        Returns:
            bool: True if the task is successful, False otherwise.
        """
        raise NotImplementedError

    @staticmethod
    @abstractmethod
    def evaluate(arena: Arena, metrics: List[str]) -> Dict[str, float]:
        """
        Evaluate the task performance and update metrics.

        Args:
            arena: The arena in which the task is being performed.
            metrics: List of metrics to evaluate.

        Returns:
            Dict[str, float]: The required evaluated metrics in a dictionary.
        """
        raise NotImplementedError

    @staticmethod
    @abstractmethod
    def reward(arena: Any) -> Dict[str, float]:
        """
        Calculate the rewards for the current state of the task.

        Args:
            arena: The arena in which the task is being performed.

        Returns:
            Dict[str, float]: The rewards for the current state of the task.
        """
        raise NotImplementedError

    @staticmethod
    @abstractmethod
    def get_goal(arena: Arena) -> InformationType:
        """
        Get the current goal of the task for its current episode.

        Args:
            arena: The arena in which the task is being performed.

        Returns:
            InformationType: The current goal of the task.
        """
        raise NotImplementedError
    
\end{lstlisting}

\newpage
\section{Action Tool} \label{sec:agent-arena-interface-actiontool}

\begin{lstlisting}[style=pythonstyle,basicstyle=\tiny\ttfamily,caption={ActionTool Interface in Agent-Arena},label={lst:action-tool}]
    
from abc import ABC, abstractmethod
from gym.spaces import Space
from typing import Dict, Any
from ..utilities.types import InformationType, ActionType, ActionSpaceType
from .arena import Arena


class ActionTool(ABC):
    """
    Abstract base class for action tools in a control problem.
    """

    def __init__(self):
        self.action_space = None

    def get_action_space(self) -> ActionSpaceType:
        """
        Get the action space of the tool.

        Returns:
            ActionSpaceType: The action space.
        """

        return self.action_space
    
    def sample_random_action(self) -> ActionType:
        """
        Sample a random action from the action space.

        Returns:
            ActionType: A randomly sampled action.
        """
        return self.action_space.sample()

    def get_action_horizon(self) -> int:
        """
        Get the action horizon (default is 0).

        Returns:
            int: The action horizon.
        """

        return 0

    def get_no_op(self) -> ActionType:
        """
        Get the no-op (no operation) action.

        Returns:
            ActionType: The no-op action.
        """
         
        raise NotImplementedError
    
    @abstractmethod
    def reset(self, arena: Arena) -> InformationType:
        """
        Reset the action tool for a new episode.

        Args:
            arena (Arena): The arena in which the action is being performed.

        Returns:
            InformationType: Information about the reset state.
        """

        raise NotImplementedError
    
    @abstractmethod
    def step(self, arena: Arena, action: ActionType) -> InformationType:
        
        """
        Perform a step in the arena using the given action.

        Args:
            arena (Arena): The arena in which the action is being performed.
            action (ActionType): The action to be taken.

        Returns:
            InformationType: Information about the result of the action.
        """

        raise NotImplementedError


\end{lstlisting}

\end{document}